\pdfoutput=1
\documentclass[11pt]{article}
\usepackage[preprint]{coling}

\usepackage{times}
\usepackage{latexsym}
\usepackage[T1]{fontenc}
\usepackage[utf8]{inputenc}
\usepackage{microtype}
\usepackage{inconsolata}
\usepackage{graphicx}
\usepackage{amsmath} 
\usepackage{arydshln}
\usepackage{multirow}

\title{SubRegWeigh: Effective and Efficient Annotation Weighing \\with Subword Regularization}

\author{
  \textbf{Kohei Tsuji\textsuperscript{1}},
  \textbf{Tatsuya Hiraoka\textsuperscript{2}},
  \textbf{Yuchang Cheng \textsuperscript{1,3}},
  \textbf{Tomoya Iwakura\textsuperscript{1,3}},
\\
\\
  \textsuperscript{1}NAIST,
  \textsuperscript{2}MBZUAI,
  \textsuperscript{3}Fujitsu Ltd.
\\
    tsuji.kohei.tl1@naist.ac.jp,\\ 
    tatsuya.hiraoka@mbzuai.ac.ae,\\ 
    \{cheng.yuchang, iwakura.tomoya\}@fujitsu.com,\\ 
}

\begin{document}
\maketitle
\begin{abstract}
NLP datasets may still contain annotation errors, even when they are manually annotated.
Researchers have attempted to develop methods to automatically reduce the adverse effect of errors in datasets.
However, existing methods are time-consuming because they require many trained models to detect errors.
This paper proposes a time-saving method that utilizes a tokenization technique called subword regularization to simulate multiple error detection models for detecting errors. 
Our proposed method, SubRegWeigh, can perform annotation weighting four to five times faster than the existing method. Additionally, SubRegWeigh improved performance in document classification and named entity recognition tasks. In experiments with pseudo-incorrect labels, SubRegWeigh clearly identifies pseudo-incorrect labels as annotation errors.
Our code is available at \url{https://github.com/4ldk/SubRegWeigh}.
\end{abstract}

\section{Introduction}
NLP datasets usually consist of raw texts and annotation labels.
Various tasks exploit the pairs of texts and labels for training and evaluating models. 
To achieve higher performance in NLP tasks, the models should be trained or fine-tuned with a sophisticated training dataset without annotation errors.

However, some popular datasets such as CoNLL-2003 contain annotation errors~\cite{CrossWeigh,identify_conll_misslabel}.
When datasets include errors, the performance is degraded by training models from incorrect training datasets, and models are incorrectly evaluated by errors of test datasets. 

Recent studies have explored automated annotation using generative AI~\cite{llm_medical_annotation,nuner} and the collaboration of humans and AI~\cite{annotation_with_ai}. 
Even with these methods, it is difficult to avoid annotation errors.
Therefore sophisticated \textbf{weighing} methods to automatically detect the annotation errors and reduce their negative effect have been expected.

Such a method to weigh annotation errors is recently studied in the NER field. 
\newcite{CrossWeigh} proposed CrossWeigh, which detects annotation errors in the dataset and adjusts their learning priority by weighting loss values so that the training is not affected by such annotation errors. 
However, there are shortcomings in its computational efficiency, especially in the recent NLP trends with the pre-trained large language models.
Reducing the computational cost contributes to not only speeding up the development of NLP but also Green AI~\cite{schwartz2019green}.
Furthermore, we cannot utilize CrossWeigh for NLP tasks other than NER because it is specially designed for NER.
Therefore, we are required to develop the new weighing methods that can be widely used for various NLP tasks. 

\begin{figure*}[t]
  \includegraphics[width=\linewidth]{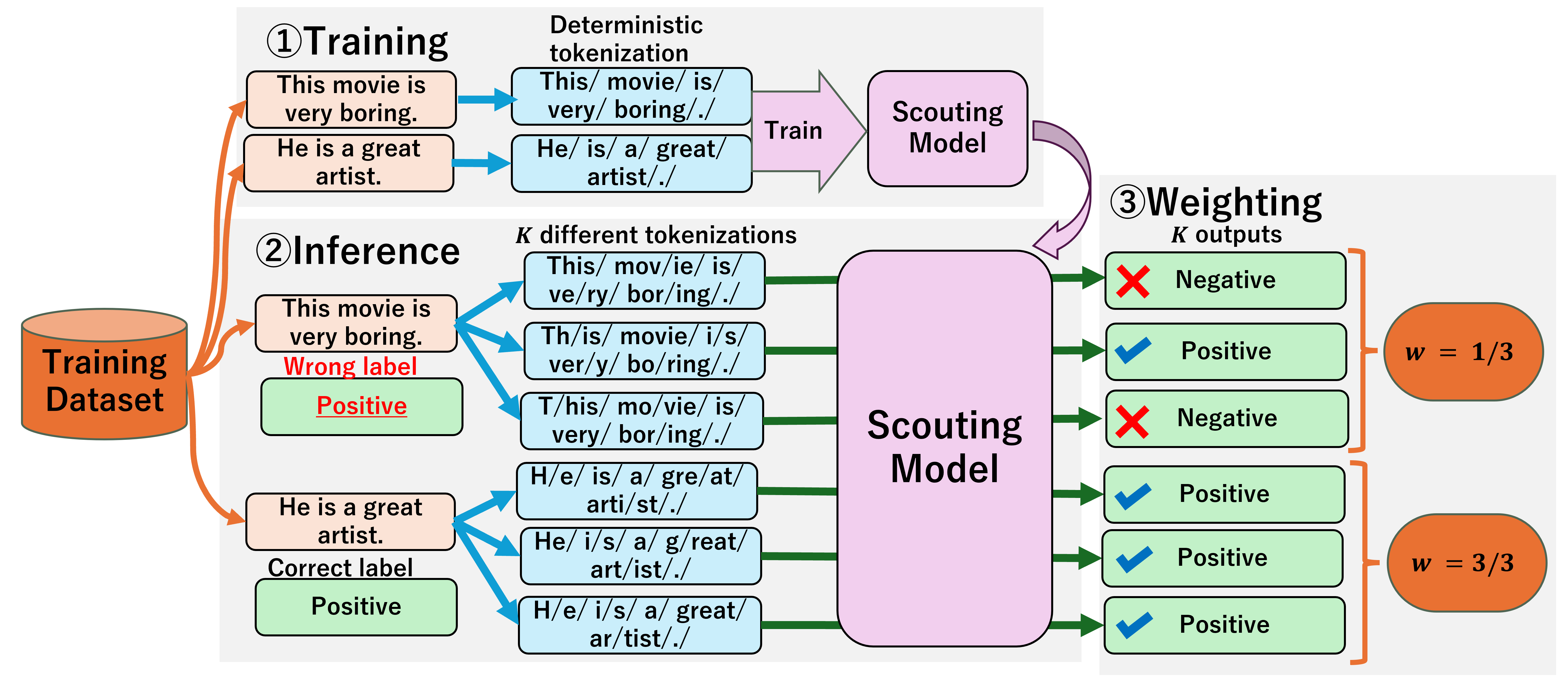}
  \caption {
    The overview of the three steps of SubRegWeigh with the number of tokenization candidates in the inference $K=3$ for the classification task whether the input sentence is positive or negative: 1) the training step of the scouting model, 2) the inference step to examine the appropriateness of the training label, and 3) the weighting step to calculate the weight of each training sample according to the appropriateness of their labels. Check marks indicate inferences that are the same as the original label and cross marks indicate inferences that differ from the original label. The calculated weights are used to train the final model. 
    }
  \label{fig:overview}
\end{figure*}

In this study, we propose an efficient and effective method of annotation weighing, \textbf{SubRegWeigh}.
Figure \ref{fig:overview} shows an overview of the proposed method.
SubRegWeigh evaluates the reliability of annotation for a single sample by feeding various input styles to a model and comparing its outputs.
Inspired by \newcite{single_model_ensemble}, we obtain such various inputs with subword regularization, which yields various tokenization candidates.
In other words, SubRegWeigh measures annotation reliability to detect errors by observing the consistency of multiple outputs generated with subword regularization.
SubRegWeigh can also be regarded as a method to simulate multiple models for annotation evaluation, which allows for faster weighing of annotation errors by using a single model instead of multiple models required by CrossWeigh.
Because the weighing processing of SubRegWeigh is not ad hoc for NER, it is applicable to wide NLP tasks.

We conducted experiments on two NLP tasks: NER and text classification.
The experimental results of the CoNLL-2003 demonstrate that SubRegWeigh can perform the annotation weighing four to five times faster than CrossWeigh.
Furthermore, SubRegWeigh contributes to the performance improvement compared to the case using CrossWeigh in some datasets. 
Especially, our proposed method achieved SoTA on CoNLL-CW, a test dataset constructed by manually correcting annotation errors in the CoNLL-2003 test split. 
Experiments with pseudo-incorrect labels also showed that the proposed method can recognize annotation errors well.

\section{Related Work}

\subsection{Annotation Error Detection}
Our work is in the line of CrossWeigh~\cite{CrossWeigh}, which is an annotation weighing method for NER.
CrossWeigh detects annotation errors by performing $K$-fold cross-validation on the training dataset $T$ times and weights the loss value according to the number of wrong predictions, which degrades the effect of training samples with annotation errors.
This method requires large computational costs because it trains $K \times T$ models for detecting annotation errors.
Our work is different from CrossWeigh in that we use multiple tokenization candidates to obtain various outputs from one model. 
This can remarkably reduce the time for detecting annotation errors because we train the model only once and use the output from pseudo-models by inputting various tokenization candidates. 
Furthermore, freed from $K$-fold cross-validation, our method can use the entire training dataset to make the model, which contributes to the more accurate detection of annotation errors.

For the case where we can prepare a small dataset with absolutely correct annotations, some methods have been studied to correct labels directly~\cite{DSNER_with_partial_annotation,ner_with_partially_annotation,partial_annotation_for_bio_ner}.
In other lines, \newcite{robust_luke} proposed a method to infer the correctness of the labels during training, which can be applied in conjunction with our method.

For detecting annotation errors in datasets beyond NER, annotation correction methods with distant supervision~\cite{DSRE} have been studied in the relation extraction tasks~\cite{DSRE_CNN, DSRE_DRL}. 
However, these methods require an additional clean dataset or are not always applicable. 
Furthermore, methods for correcting POS tagging tasks have also been studied~\cite{pos_correct1, pos_correct2}, but these do not apply to tasks other than POS tagging. 
The analyzing method of the models' memorization~\cite{example_tied_dropout} could be used to detect training samples with ambiguous labels.

\subsection{Subword Regularization}
The proposed method exploits the technique of tokenization for efficient annotation error detection.
Tokenization~\cite{word_piece, bpe, subword_reg} is widely used in recent NLP. 
This technique reduces the number of unknown and low-frequency words and limits the number of dictionaries by treating words as a sequence of finer subwords. 
However, because tokenization is deterministically performed on training data, the subword sequences of inference data may not always match the sequences learned during training.

Subword regularization~\cite{subword_reg, bpe_dropout, mmt} alleviates this problem of deterministic tokenization.
This technique generates multiple tokenization candidates for a single text. 
While subword regularization is typically used during the training step to learn from various tokenizations, some studies use it during inference. 
\newcite{single_model_ensemble} uses subword regularization during inference to ensemble the various outputs of a single model by inputting some different tokenization candidates to the model.
\newcite{multi_view} proposed a method to reduce the effect of tokenization difference. 
However, there has been no research using subword regularization during inference to reduce noise in the training data.

Subword regularization using a unigram language model~\cite{subword_reg} is a method to sample tokenization candidates using the unigram probability of tokens.
For the cases where the model does not employ a tokenizer with the unigram language model like Byte Pair Encoding (BPE)~\cite{bpe} used in RoBERTa~\cite{roberta} or WordPiece~\cite{word_piece} used in BERT~\cite{bert}, we can use BPE-Dropout~\cite{bpe_dropout} and MaxMatch-Dropout~\cite{mmt} for the subword regularization, respectively.
To obtain various tokenization candidates, BPE-Dropout randomly avoids merges during the combination process and MaxMatch-Dropout randomly rejects the longest match candidates for each word.

\section{Proposed Method: SubRegWeigh}
The purpose of the proposed method, SubRegWeigh, is to detect the incorrectly annotated samples in the training dataset and reduce the effect of such a sample with annotation errors during the training of the final model by weighting the loss values. 
The process of SubRegWeigh is composed of the following three steps (Figure~\ref{fig:overview}):
\begin{enumerate}
\setlength{\parskip}{0.0cm}
\setlength{\itemsep}{0.1cm}
\item Training \textbf{a scouting model} in a usual manner of NLP tasks with the training data using deterministic tokenization (\S \ref{sec:training}).
\item Inferring the labels for the training samples with the trained scouting model. Herein, we tokenize the input texts into $K$ different tokenization candidates. Therefore, we can collect $K$ inferred outputs for each tokenization candidate of a single input text (\S \ref{sec:inference}).
\item Calculating weights for training samples according to the number of correctly answered inputs for each tokenization candidate (\S \ref{sec:weighing}).
\end{enumerate}
The first and second steps detect the annotation errors by scouting the dataset. 
Then, SubRegWeigh weighs the correctness of annotation labels and assigns the weight for each training sample in the third step.
After the above steps, we can train \textbf{the final model} with the original training data and the calculated weights for each training sample (\S \ref{sec:finalizing}).
Compared to CrossWeigh, SubRegWeigh can be used for various NLP tasks beyond NER because the entire steps are not specialized to NER.
The following section explains each step in detail.

\subsection{Training of the Scouting Model}
\label{sec:training}
Inspired by CossWeigh~\cite{CrossWeigh}, we first create the scouting model $M$ (see the left top of Figure \ref{fig:overview}).
This model is used to scout the correctness of the annotation in the training dataset in the next step.
Let $(X, Y)$ be a pair of the text and the label corresponding to a single training sample. 
When the text is composed of $I$ words, $X = x_1, ... , x_I$. 
$Y$ is a single value in text classification (e.g., positive and negative labels) or a sequence of labels $Y = y_1, ... , y_I$ in sequential labeling such as NER. 

In this step, we train the scouting model $M$ in the usual manner of training or fine-tuning for general NLP datasets.
The single input text $X$ is tokenized into a sequence of subwords $X'=s_1, ..., s_J$ that is composed of $J$ subwords\footnote{The numbers of words $I$ and subwords $J$ are not necessarily matched depending on the tokenization methods. In the case of sequential labeling tasks with $I \neq J$, we aligned the subwords and labels following the existing literature~\cite{roberta, luke}.}.
Then, we train the scouting model $M$ with the tokenized texts and labels.
Note that we use the deterministic tokenization here (e.g., the default encoding of the BPE tokenizer or the $1$-best tokenization with the Viterbi algorithm for the unigram language model-based tokenizer).
In other words, we do not use subword regularization (i.e., stochastic tokenization) for the training of the scouting model $M$.

\subsection{Inference with the Scouting Model}
\label{sec:inference}
In the second step, we use the trained scouting model $M$ to scout the correctness of the annotated label in the training dataset (see the left bottom of Figure \ref{fig:overview}).
We assume that the trained scouting model $M$ can correctly predict the labels even for the differently tokenized input if the label is correctly annotated.
Concretely, the input text $X$ is tokenized into $K$ different tokenization candidates $\{X'_1, ..., X'_k, ..., X'_K\}$. 
Then each tokenization candidate $X'_k$ is fed into the trained model $M$ and we obtain the corresponding prediction $\hat{Y}_k$\footnote{As well as $Y$, $\hat{Y}_k$ can be single or multiple labels depending on target tasks}.
We then evaluate the output $\hat{Y}_k$ with the gold annotation $Y$.
This process can be regarded as the ``mistake-reweighing'' process of CrossWeigh with $K$ pseudo-models, which is realized with the $K$ different tokenization candidates for the single input text and the single model~\cite{single_model_ensemble}.

We can obtain various tokenization candidates using a technique of subword regularization.
When we use BPE or WordPiece, BPE-Dropout and MaxMatch-Dropout can be used, respectively.
When one uses a unigram language model-based tokenizer, we can also use $N$-best tokenization instead of subword regularization.

We consider it important to make a variety of the $K$ tokenization candidates to weigh the correctness of the annotation from diverse aspects.
To avoid using similar subword sequences as the inputs for the model $M$, we select $K$ more different candidates from a large number $N$ of tokenization candidates.
We propose three options of the way to select such $K$ tokenization candidates as follows.

\paragraph{Random:} 
We randomly sample $N=K$ different tokenization candidates using subword regularization and use them as inputs for the model $M$.
In other words, we do not select the tokenization candidates from the large number of candidates.

\paragraph{Cos-Sim:} 
We select $K$ tokenization candidates using the cosine similarity calculated with TF-IDF vectors. TF-IDF was calculated with each subword as a term, each subword sequence as a document, and the entire candidate set as a corpus.
First, we convert $N (> K$) tokenization candidates into TF-IDF vectors and select the first candidate with the smallest cosine similarity against the deterministic tokenization $X'$.
Subsequently, we select the candidate that is least similar to both $X'$ and already selected candidates.
We repeatedly select the candidates until the number of candidates reaches $K$.

\paragraph{$K$-means:}
We select $K$ tokenization candidates using the $K$-means clustering.
We apply the clustering to $N (> K)$ candidates vectorized with TF-IDF.
We then choose $K$ candidates at the nearest point to the centroids of each cluster.

Table \ref{tab:tokenization_examples} in Appendix \ref{sec:tokenization_examples} shows the tokenization examples by each selection method.

\subsection{Weighting for Training Samples}
\label{sec:weighing}
To reduce the effect of training samples with annotation errors, we calculate weights for each sample depending on the results of the inference step (see the right part of Figure \ref{fig:overview}).
Concretely, we calculate the weight $w$ corresponding to the training sample $(X, Y)$ as the follows:
\begin{align}
w &= \min(w_{\mathrm{min}}, \frac{1}{K}\sum_{k=1}^{K}{c}_{k}), \label{eq:weight}\\
    {c}_{k} &= 
   \begin{cases}
      1 & \text{if} \quad \hat{Y}_k = Y \\
      0 & \text{otherwise}
   \end{cases},
\end{align}
where $\hat{Y}_k$ is the output of the scouting model $M$ when inputting the tokenization candidate $X'_k$.
If all predictions were different from the original label $Y$, the weights would be $0$ and the data would not be used for training. 
We use a pre-defined minimum weight $w_\mathrm{min}$ to avoid the zero weights, following the suggestion in \newcite{CrossWeigh} finding it better to give a small weight to samples with annotation errors than not to use them.

\subsection{Training of Final Model} 
\label{sec:finalizing}
Using the training dataset with weights calculated in the above section, we train the final model $M'$ that is independent of $M$.
Following CrossWeigh~\cite{CrossWeigh}, the loss value $L_{\mathrm{weighted}}$ for the training sample $(X, Y)$ with the weight $w$ is defined as the follows:
\begin{equation}
L_{\mathrm{weighted}}(X, Y, w) = w L_{\mathrm{original}}(X, Y),
\end{equation}
where $L_{\mathrm{original}}(\cdot)$ is the loss function used in the default training.

\section{Experiments}
\label{sec:experiment}
We examine the processing time for the annotation error detection and weighting.
Besides, we evaluate the performance of NER and text classification when using the training dataset with the weights.

\subsection{Dataset} 
We briefly overview the dataset used in the experiments here. 
Appendix~\ref{sec:detail_dataset} also explains in detail.
\paragraph{NER:}
We used \textbf{CoNLL-2003}~\cite{conll2003} with the official split for the training, validation, and test. The label format was changed from IOB1 to IOB2~\cite{iob2}.
In addition, we used \textbf{CoNLL-CW}~\cite{CrossWeigh} and \textbf{CoNLL-2020}~\cite{conll2023} for the test dataset.\footnote{Because both of these dataset names are CoNLL++, we call CoNLL-CW and CoNLL-2020 to distinguish them.}

\paragraph{Text Classification:}
We used \textbf{SST-2}~\cite{sst2}.
The official training split was used for the training dataset and the validation split was used for the test dataset.

\subsection{Compared Methods}
We compared the following five methods including three options of SubRegWeigh explained in \S \ref{sec:inference}.

For the baselines, we selected \textbf{Vanilla} and \textbf{CrossWeigh}.
\textbf{Vanilla} is the final model that is trained on the original training split without any annotation weighing.
\textbf{CrossWeigh} is the final model that is trained with the dataset weighed by the official code\footnote{\url{https://github.com/ZihanWangKi/CrossWeigh}} of the existing work \cite{CrossWeigh}. 
Although CrossWeigh is specialized to NER, we forcibly utilize it for the text classification setting without Entity Disjoint~\cite{CrossWeigh}.

We evaluated the three options of our method introduced in \S \ref{sec:inference}, which are represented as \textbf{SubRegWeigh (Random)}, \textbf{SubRegWeigh (Cos-Sim)}, and \textbf{SubRegWeigh ($K$-means)}, respectively.
We used these proposed methods to weight the training dataset and evaluate the final model trained with this weighted dataset as well as CrossWeigh.

\subsection{Model Settings}
\label{sec:model_settings}
We employed $\mathrm{RoBERTa_{LARGE}}$ as the pre-trained language model for both NER and text classification.
Besides $\mathrm{LUKE_{LARGE}}$~\cite{luke} is used for the NER setting because LUKE is an architecture for entity-related tasks.
In this experimental setting, we used the same backbone model for both the scouting model and the final model\footnote{\S\ref{sec:different_backbone} discusses the case where using different pre-trained models between the scouting and final models.}.
For example, when RoBERTa is used in the scouting model, the final model will also use RoBERTa.
The detailed model settings such as hyperparameters are provided in the Appendix~\ref{sec:detail_model_settings}.

We set the three hyperparameters of CrossWeigh: the number of folds in mistake estimation $K=10$, the number of iterations of mistake estimation $T=3$, and the weight scaling factor $\epsilon = 0.7$.
BPE-Dropout~\cite{bpe_dropout} was used to obtain the multiple tokenization candidates for SubRegWeigh because both RoBERTa and LUKE employ BPE for their tokenizers. 
BPE-Dropout has a hyperparameter $p$, where a higher $p$ results in finer tokenization (i.e., more different from the original tokenization). 
In the experiments, $p$ was set to $0.1$ unless otherwise noted. 
Additionally, by default, the hyperparameters for SubRegWeigh were set to $N=500$, $K=10$, and $w_{\mathrm{min}}=1/3$.

\subsection{Evaluation Metrics} 
We evaluate the methods from the viewpoints of speed and performance.
On the speed side, we measured the time for detecting the annotation errors and generating the weighted data for the training dataset in NER\footnote{We measured the processing time with Xeon Platinum 8167M + NVIDIA V100 for RoBERTa and Xeon Gold 6230R + NVIDIA A100 for LUKE.}.
On the performance side, we measured the $F_1$ score in NER and the accuracy in text classification on each test dataset for the final models trained on the weighted training data.

\begin{table*}[t]
\centering
\small
\begin{tabular}{lrrrrrr}
\hline \hline
 & \multicolumn{1}{c}{\begin{tabular}[c]{@{}c@{}}Processing\\ time\end{tabular}} & \multicolumn{1}{c}{\begin{tabular}[c]{@{}c@{}}CoNLL\\ valid\end{tabular}} & \multicolumn{1}{c}{\begin{tabular}[c]{@{}c@{}}CoNLL\\ test\end{tabular}} & \multicolumn{1}{c}{\begin{tabular}[c]{@{}c@{}}CoNLL\\ CW\end{tabular}} & \multicolumn{1}{c}{\begin{tabular}[c]{@{}c@{}}CoNLL\\ 2020\end{tabular}} & \multicolumn{1}{c}{SST-2} \\
 & \multicolumn{1}{c}{hh:mm} & \multicolumn{1}{c}{\(F_1\)} & \multicolumn{1}{c}{\(F_1\)} & \multicolumn{1}{c}{\(F_1\)} & \multicolumn{1}{c}{\(F_1\)} & \multicolumn{1}{c}{ACC} \\ \hline
SoTA & \multicolumn{1}{c}{-} & \multicolumn{1}{c}{-} & ${94.60}^{\dagger}$ & ${95.88}^{\dagger\dagger}$ & \multicolumn{1}{c}{-} & \multicolumn{1}{c}{-} \\ \hline
$\mathrm{RoBERTa_{LARGE}}$ & \multicolumn{1}{l}{} & \multicolumn{1}{c}{} &  &  &  &  \\
Vanilla & \multicolumn{1}{c}{-} & 97.26${}_{\pm0.12}$ & 93.54${}_{\pm0.28}$ & 95.27${}_{\pm0.24}$ & 94.80${}_{\pm0.22}$ & 94.68${}_{\pm0.12}$ \\
CrossWeigh & 30:55 & 97.11${}_{\pm0.11}$ & 93.40${}_{\pm0.21}$ & 94.99${}_{\pm0.19}$ & 94.93${}_{\pm0.25}$ & 94.31${}_{\pm0.09}$ \\
SubRegWeigh & \multicolumn{1}{l}{} & \multicolumn{1}{l}{} & \multicolumn{1}{l}{} & \multicolumn{1}{l}{} & \multicolumn{1}{l}{} &  \\
\multicolumn{1}{r}{Random} & 3:26 & \textbf{97.30${}_{\pm0.16}$} & 93.51${}_{\pm0.26}$ & 95.24${}_{\pm0.22}$ & 94.52${}_{\pm0.22}$ & 94.61${}_{\pm0.10}$ \\
\multicolumn{1}{r}{Cos-Sim} & 4:51 & 97.27${}_{\pm0.12}$ & 93.44${}_{\pm0.21}$ & 95.17${}_{\pm0.23}$ & 94.91${}_{\pm0.19}$ & 94.75${}_{\pm0.09}$ \\
\multicolumn{1}{r}{$K$-means} & 5:21 & 97.28${}_{\pm0.11}$ & \textbf{93.81${}_{\pm0.16}$} & \textbf{95.45${}_{\pm0.25}$} & \textbf{94.96${}_{\pm0.21}$} & \textbf{94.84${}_{\pm0.07}$} \\ \hline
$\mathrm{LUKE_{LARGE}}$ & \multicolumn{1}{c}{} &  &  &  &  &  \\
Vanilla & \multicolumn{1}{c}{-} & \textbf{96.78${}_{\pm0.08}$} & \textbf{94.32${}_{\pm.0.15}$} & 95.92${}_{\pm0.13}$ & 95.29${}_{\pm0.18}$ & \multicolumn{1}{c}{-} \\
CrossWeigh & 26:19 & 96.62${}_{\pm0.08}$ & 94.12${}_{\pm.0.19}$ & 95.96${}_{\pm0.12}$ & \textbf{95.32${}_{\pm0.18}$} & \multicolumn{1}{c}{-} \\
SubRegWeigh &  & \multicolumn{1}{l}{} & \multicolumn{1}{l}{} & \multicolumn{1}{l}{} & \multicolumn{1}{l}{} &  \\
\multicolumn{1}{r}{Random} & 2:59 & 96.54${}_{\pm0.11}$ & 94.22${}_{\pm0.13}$ & 95.94${}_{\pm0.15}$ & 95.24${}_{\pm0.20}$ & \multicolumn{1}{c}{-} \\
\multicolumn{1}{r}{Cos-Sim} & 6:19 & 96.53${}_{\pm0.10}$ & 94.11${}_{\pm0.16}$ & 95.93${}_{\pm0.13}$ & 95.21${}_{\pm0.25}$ & \multicolumn{1}{c}{-} \\
\multicolumn{1}{r}{$K$-means} & 6:36 & 96.65${}_{\pm0.09}$ & 94.20${}_{\pm0.15}$ & \textbf{96.12${}_{\pm0.18}$} & 95.31${}_{\pm0.14}$ & \multicolumn{1}{c}{-} \\ \hline \hline
\end{tabular}
\caption{
    The processing time and the performance of the final models with the weighted dataset. 
    The best performing scores are highlighted in bold.
    ${}^\dagger$ and ${}^{\dagger\dagger}$ is SoTA score from \newcite{ace} and ~\newcite{robust_luke}.
}
\label{tab:main_results}
\end{table*}

\subsection{Experimental Results}
The results of our experiments are shown in Table~\ref{tab:main_results}.
We reported the averaged scores and the standard deviations over the five independent runs.

\subsubsection{Processing Time}
The column ``Processing time'' shows the time taken to detect the annotation errors and assign weights for each sample in the training dataset in NER.
As shown in this column, the proposed method, SubRegWeigh (Random), was maximum of over eight times faster than CrossWeigh.
Even the most time-consuming method of SubRegWeigh ($K$-means) was five times faster with RoBERTa and four times faster with LUKE than CrossWeigh. 
The method with the random selection is the fastest among the proposed methods because it does not require any of the calculations about TF-IDF, cosine similarity, and $K$-means clustering.
In addition, in the random method, only $K(=N=10)$ tokenization candidates were generated, whereas the other methods of SubRegWeigh need to select $K$ candidates from the $N(=500)$-sampled tokenization candidates, which results in longer processing time.
Although the proposed method of Cos-Sim and $K$-means takes longer time than Random, it is remarkably faster than the existing method for the automatic annotation weighing, CrossWeigh.

\subsubsection{Performance on NER}
The performance on CoNLL-CW is the most important in our evaluation because annotation errors in the test split are removed, which indicates that we can compare the methods on the most unpolluted dataset\footnote{The vanilla baseline score of LUKE exceeds the SoTA score~\cite{robust_luke} for CoNLL-CW, which we consider because of the successful hyperparameter setting.}. 
One can see that SubRegWeigh ($K$-means) achieves higher performance than the vanilla method on the test dataset of CoNLL-CW.
Specifically, SubRegWeigh ($K$-means) improved the $F_1$ scores by 0.18 points with RoBERTa and 0.20 points with LUKE compared to each vanilla baseline.
This result indicates that the proposed method of annotation weighing contributes to the performance improvement in the clean test dataset.
Furthermore, SubRegWeigh ($K$-means) achieves higher scores than CrossWeigh, which demonstrates that the proposed method is a reasonable alternative to the existing method.

Both SubRegWeigh and CrossWeigh scored lower than Vanilla in the CoNLL valid and test datasets, which contain annotation errors. 
These results indicate the remarkable negative effect of the annotation errors in the valid and test datasets and the difficulty of evaluating the models appropriately.

Among the proposed methods, Random scores lower than $K$-means in many cases.
We consider this because the random selection of tokenization candidates leads to the selection of similar subword sequences, causing a bias in the inference results.
Cos-Sim also scored lower than $K$-means in all cases, which shows that the naive method of selecting tokenization candidates does not contribute to performance improvement.
We consider that the $K$-means clustering can handle the diverse range of subword tokenization candidates, and this leads to the successful result.

\subsubsection{Performance on Text Classification}
For text classification, SubRegWeigh ($K$-means) achieved the highest accuracy, and SubRegWeigh (random) had lower accuracy, similar to the results in NER. 
One reason for the lack of performance improvement with CrossWeigh in SST-2 is that Entity Disjoint is important for CrossWeigh but unavailable in the text classification task.

From the results on the NER and text classification datasets, we conclude that the proposed method is superior to the existing method in terms of processing time and task performance.
Especially, we can say SubRegWeigh ($K$-means) is the best option, which achieves the highest task performance with approximately four to five times faster than CrossWeigh.

\section{Discussion}

\subsection{Pseudo-incorrect Label Test}
\label{sec:pseudo_incorrect_label_test}
\begin{table}
\small
\centering
\begin{tabular}{lcccc}
\hline \hline
 & $\overline{w}_{\mathrm{cor}}$ & $\overline{w}_{\mathrm{incor}}$ & \multicolumn{1}{c}{\begin{tabular}[c]{@{}c@{}}CoNLL\\ test\end{tabular}} & \multicolumn{1}{c}{\begin{tabular}[c]{@{}c@{}}CoNLL\\ CW\end{tabular}} \\
 & \multicolumn{1}{c}{} & \multicolumn{1}{c}{} & \multicolumn{1}{c}{\(F_1\)} & \multicolumn{1}{c}{\(F_1\)} \\ \hline
Vanilla & \multicolumn{1}{c}{-} & \multicolumn{1}{c}{-} & 84.15 & 85.49 \\
CrossWeigh & 0.8000 & 0.0019 & 84.28 & 85.71 \\
SubRegWeigh &  &  &  &  \\
\multicolumn{1}{r}{Random} & 0.9278 & 0.0048 & 83.77 & 85.04 \\
\multicolumn{1}{r}{Cos-Sim} & 0.8346 & 0.0042 & 84.25 & 85.69 \\
\multicolumn{1}{r}{$K$-means} & 0.9284 & 0.0048 & 84.34 & 85.76 \\
\hline \hline
\end{tabular}
\caption{Averaged weights and performance ($F_1$) for the pseudo-incorrect dataset. $\overline{w}_{\mathrm{cor}}$ and $\overline{w}_{\mathrm{incor}}$ are the averaged weights for the samples with original labels and ones with the pseudo-incorrect labels, respectively.}
\label{tab:pseudo_incorrect_label}
\end{table}

We verified whether SubRegWeigh can accurately detect annotation errors in the dataset.
We evaluate this capability using a modified dataset with some labels flipped to incorrect labels artificially.

Assuming that the original label should be correctly annotated, we replaced 10\% of the labels in the CoNLL2003 training dataset with different labels that do not conflict with the IOB2 format (Appendix \ref{sec:pseudo_incorrect_label_creation}). 
Then, 3,329 sentences including pseudo-incorrect labels, and 5,356 sentences with original annotations are obtained.

We assign weights to this modified dataset using CrossWeigh and SubRegWeigh\footnote{The hyperparameter settings are the same as those described in Section \S \ref{sec:model_settings} and Appendix \ref{sec:detail_model_settings}.}.
Then we compared the averaged weights between the pseudo-incorrect and the original samples to confirm whether annotation error detection is correct. 
Here, we calculate the weights as $\frac{C}{K}$, where $C$ is the number of inferences that the model predicts the original label, instead of $w$ in Eq. \eqref{eq:weight}.
This is because we consider that the minimum weight $w_{min}$ and the weight correction in CrossWeigh could have a negative effect on the fair analysis.

In Table \ref{tab:pseudo_incorrect_label}, $\overline{w}_{\mathrm{incor}}$ is the averaged weights assigned to the samples with pseudo-incorrect labels, where we expect them should be lower as incorrect labels.
Similarly, $\overline{w}_{\mathrm{cor}}$ is the averaged weights assigned to the original labels, where higher weights should be assigned as the correct labels.

From the results in this table, SubRegWeigh ($K$-means) assigns the most distinct weights between the replaced and the unreplaced data. 
Additionally, all methods assign weights approximately 100 times lower to replaced data than unreplaced data. 
This indicates that all methods can detect errors effectively. 
The lowest average weight for unreplaced data is CrossWeigh. 
This is because CrossWeigh infers only 3 times while SubRegWeigh infers 10 times. 
Even one incorrect prediction reduced the weight to 2/3. 
Among SubRegWeigh methods, Cos-Sim assigns lower weights to sentences without pseudo-incorrect labels, but these do not mean lower performance improvements compared to other methods, especially SubRegWeigh (Random). 
According to this, if the weights of sentences containing errors are low, it does not matter if the average weight is slightly lower.

In addition to the investigation of weights, we also analyze the obtained performance by training final models with the modified dataset.
For the training of the final models, we used the weights calculated in the manner of Eq. \eqref{eq:weight}.
The results are shown in the right two columns of Table \ref{tab:pseudo_incorrect_label}.
Because the training data is noisy, the entire scores are worse than the ones in Table \ref{tab:main_results}.
However, the total tendency of the performance is similar to the results in \S \ref{sec:experiment}.
From the entire results, we cannot find a clear relationship between the averaged weights and the final performance.

\subsection{Numbers of Tokenization Candidates}

\begin{table}[t]
\small
\centering
\begin{tabular}{rrrrr}
\hline \hline
\multicolumn{1}{c}{$K$}   & \multicolumn{1}{c}{Method} & Time &  \multicolumn{1}{c}{\begin{tabular}[c]{@{}c@{}}CoNLL\\ test\end{tabular}} & \multicolumn{1}{c}{\begin{tabular}[c]{@{}c@{}}CoNLL\\ CW\end{tabular}} \\ \hline
500 & Random & 75:41 & 93.71 & 95.26\\\hdashline
50  & Random & 9:15  & 93.46 & 95.08\\
50  & Cos-sim& 10:52 & 93.46 & 95.12\\
50  & $K$-means& 12:02 & 93.65 & 95.30\\\hdashline
10  & Random & 3:26  & 93.51 & 95.24\\
10  & Cos-Sim& 4:51  & 93.44 & 95.17\\
10  & $K$-means& 5:21  & 93.81 & 95.45\\ \hline \hline
\end{tabular}
\caption{Difference in the processing time (hh:mm) and the performance ($F_1$) against three options of $K$.}
\label{tab:number_of_sub_seq}
\end{table}

\begin{table}[t]
\centering
\small
\begin{tabular}{llrr}
\hline \hline
\begin{tabular}[c]
{@{}c@{}}Final \\ Model\end{tabular} & \begin{tabular}[c]{@{}c@{}}Scouting\\ Model\end{tabular} & \multicolumn{1}{l}{\begin{tabular}[c]{@{}c@{}}CoNLL\\ test\end{tabular}} & \multicolumn{1}{l}{\begin{tabular}[c]{@{}c@{}}CoNLL\\ CW\end{tabular}} \\ \hline
RoBERTa   & RoBERTa    & 93.81        & 95.45       \\
          & LUKE       & 93.65        & 95.42       \\ \hdashline
LUKE      & RoBERTa    & 94.24        & 95.89       \\
          & LUKE       & 94.20        & 96.12       \\ \hline \hline
\end{tabular}
\caption{Difference in the performance ($F_1$) when using different pre-trained models for scouting and final model.}
\label{tab:different_model}
\end{table}

\begin{table*}[t!]
\small
\centering
\begin{tabular}{l|cccc}
\hline \hline
\multicolumn{1}{c|}{\multirow{2}{*}{Text${}_{\mathrm{Label}}$}}                                                                                                                                                                                                                                                                                                                                                                                                                                                   & \multicolumn{1}{c}{\multirow{2}{*}{CrossWeigh}} & \multicolumn{3}{c}{SubRegWeigh}                                                        \\
\multicolumn{1}{c|}{}                                                                                                                                                                                                                                                                                                                                                                                                                                                                                             & \multicolumn{1}{c}{}                            & \multicolumn{1}{c}{Random} & \multicolumn{1}{c}{Cos-Sim} & \multicolumn{1}{c}{K-means} \\ \hline
\begin{tabular}[c]{@{}l@{}}The${}_{\mathrm{O}}$ foreign${}_{\mathrm{O}}$ ministry${}_{\mathrm{O}}$ ’s${}_{\mathrm{O}}$ Shen${}_{\underline{\mathrm{B-ORG}}}$ told${}_{\mathrm{O}}$ \\ Reuters${}_{\mathrm{B-ORG}}$ Television${}_{\mathrm{I-ORG}}$ in${}_{\mathrm{O}}$ an${}_{\mathrm{O}}$ interview${}_{\mathrm{O}}$ \\ he${}_{\mathrm{O}}$ had${}_{\mathrm{O}}$ read${}_{\mathrm{O}}$ reports${}_{\mathrm{O}}$ of${}_{\mathrm{O}}$ Tang${}_{\mathrm{B-PER}}$ ’s${}_{\mathrm{O}}$ comments${}_{\mathrm{O}}$ ...\end{tabular} & 0.343                                           & 0.333                      & 0.333                       & 0.333                       \\ \hdashline
\begin{tabular}[c]{@{}l@{}}NOTES${}_{\mathrm{O}}$ BAYERISCHE${}_{\mathrm{B-ORG}}$ VEREINSBANK${}_{\mathrm{I-ORG}}$ IS${}_{\mathrm{O}}$ \\ JOINT${}_{\mathrm{O}}$ LEAD${}_{\mathrm{O}}$ MANAGER${}_{\mathrm{O}}$\end{tabular}                                                                                                                                                                                                                                                                                      & 0.700                                           & 0.500                      & 0.500                       & 0.333                       \\ \hdashline
\begin{tabular}[c]{@{}l@{}}Former${}_{\mathrm{O}}$ Surinam${}_{\underline{\mathrm{B-LOC}}}$ rebel${}_{\underline{\mathrm{O}}}$ \\ leader${}_{\mathrm{O}}$ held${}_{\mathrm{O}}$ after${}_{\mathrm{O}}$ shooting${}_{\mathrm{O}}$ .${}_{\mathrm{O}}$\end{tabular}                                                                                                                                                                                                                                                                          & 1.000                                           & 0.700                      & 0.700                       & 0.500                       \\ \hline \hline
\end{tabular}
\caption{Examples of weights assigned by each method. Underlined are incorrect or ambiguous labels.}
\label{tab:case_study}
\end{table*}
$K$ is an important hyperparameter affecting both processing time and performance of SubRegWeigh. 
We investigated the difference in the processing time and the $F_1$ score of the final model with three options of the number of tokenization candidates $K$ in the NER dataset. 
We examined the difference in the range of $K={10, 50, 500}$ with RoBERTa.
The dropout rate for BPE-Dropout was $p=0.1$.
Since $K$-means and Cos-Sim are techniques to reduce the variations in subword segmentation, a large value like $K=500$ makes them almost indistinguishable from the Random method.
Therefore, only the Random method was investigated for $K=500$.

The results are shown in Table~\ref{tab:number_of_sub_seq}. 
Comparing the results between the ones with $K=10$ and $K=50$, one can see that $K$ does not have a large effect on the $F_1$ scores.
However, the larger $K$ remarkably increases the processing time.
For $K=500$, the $F_1$ score improved compared to Random with $K=10$ and $K=50$ on the CoNLL and CoNLL-CW datasets. 
This suggests that using a large number of subword sequences can effectively weigh annotation errors. 
However, such much large $K$ damages the speed of annotation weighing, making it a trade-off between the speed vs. the performance.
While increasing \(K\) to 500 in the Random method showed performance improvement, $K$-means exhibited the highest performance even with smaller \(K\). 
This indicates that $K$-means can select sufficiently diverse subword sequences.

\subsection{Weighting with Different Backbone}
\label{sec:different_backbone}
In the experiment shown in \S \ref{sec:experiment}, we used the same backbone for the scouting model $M$ for annotation weighing and the final model $M'$.
Herein, we investigated the performance difference when using different backbones in NER.
Specifically, we examined the effect in \(F_1\) scores when $M$ was RoBERTa and $M'$ was LUKE.
Similarly, we examined the case where $M$ was LUKE and $M'$ was RoBERTa.
Both models were trained with the hyperparameters explained in Table \ref{tab:hyper_param}.
We used $K$-means for selecting the tokenization candidates because it showed the best performance in the main results (\S \ref{sec:experiment}).

The results in Table~\ref{tab:different_model} indicate that, in most cases, the data weighted by the same model tends to achieve higher \(f_1\) scores. 
Although we can save time by preparing the weighed dataset with a single model and reusing it to train models with other backbones, this observation suggests that we should prepare the weighed dataset depending on the backbone used for the final models to obtain performance improvement.
This suggestion also supports the importance of the fast method of annotation weighing.

\subsection{Qualitative Analysis}
\label{sec:qualitative_analysis}

Several examples of the CoNLL-2003 training dataset weighed by each method are shown in Table~\ref{tab:case_study}. 
All methods successfully assigned low weights for clearly incorrect sentences like the top example in the table. 
However, for ambiguous labels like the example at the bottom of the table, CrossWeigh tended to assign high weights, whereas SubRegWeigh more frequently assigned low weights. 
Additionally, we discovered a specific weakness of SubRegWeigh: it tends to assign lower weights to sentences composed entirely of uppercase letters (see the example in the middle of the table). 
This is because only a few uppercase-only subwords are included in the tokenizer's vocabulary, and even with a low \(p\), subword regularization significantly changes the tokenization candidates for uppercase-only sentences, leading to incorrect inferences.
We believe this issue is not unique to our method but rather a general problem with subword regularization, which we plan to investigate further in future research.

\section{Conclusion} 

We proposed SubRegWeigh, a method for annotation weighing using subword regularization, which offers faster annotation weighing than existing methods. 
In particular, subword sequence selection using $K$-means was four to five times faster than CrossWeigh for annotation weighing and contributed to better performance than weighting with all of the large number of subword sequences.

In addition, the performance dropped in many cases when different models were used for the scouting and final model, indicating the need for comparison including annotation weighing for a better model and the importance of developing an efficient annotation weighing method.

\section*{Limitation}
In this paper, we proposed a method for annotation weighing using subword regularization. However, since a deep learning model is used for error weighing, the calculated weights are not always guaranteed to be appropriate. 

As mentioned in \ref{sec:qualitative_analysis}, if there is not enough variation of the subword sequence, such as uppercase-only sentences, our proposed method often assigns incorrect weights.

In \S \ref{sec:pseudo_incorrect_label_test}, we created pseudo-incorrect labels by randomly replacing the original labels with incorrect labels. This error distribution may differ from real-world annotation error distribution. 

\section*{Ethical Considerations}
Experiments presented in this work were performed with existing datasets~\cite{conll2003, CrossWeigh, conll2023, sst2}. These datasets were used to study NER or text classification models, which is consistent with their intended use. Because this study focuses on efficient annotation error detection in the dataset, its potential risks and negative impact on society appear to be minimal.

\bibliography{main}

\appendix
\section{Example For Each  Selection Method}
\label{sec:tokenization_examples}
\begin{table*}[t]
\centering
\small
\begin{tabular}{llc}
\hline \hline
Method & Subwords & \multicolumn{1}{c}{\begin{tabular}[c]{@{}c@{}}the number of\\ subwords\end{tabular}} \\ \hline
Default Tokenization & \begin{tabular}[c]{@{}l@{}}Japan then laid siege to the Syrian penalty area for most of the game \\ but rarely breached the Syrian defence .\end{tabular} & 20 \\ \hline
Random & \begin{tabular}[c]{@{}l@{}}Japan then la id s iege to t he Sy rian penalty ar ea for most of the game \\ bu t rarely bre ache d t he Sy rian def ence .\end{tabular} & 29 \\ \cdashline{2-3}
 & \begin{tabular}[c]{@{}l@{}}J a pan then l ai d s ieg e to the Syrian penal ty a rea for most of the ga me \\ bu t rarely bre ache d th e Syrian de fen ce .\end{tabular} & 33 \\ \cdashline{2-3}
 & \begin{tabular}[c]{@{}l@{}}Japan then laid siege to t he Syrian pen alt y ar ea f o r mo st of the ga me \\ but rarely b rea ched the Sy r ian defence .\end{tabular} & 30 \\ \hline
Cos-Sim & \begin{tabular}[c]{@{}l@{}}J ap an t hen l ai d s ie ge to th e S y rian p en alty a rea for most o f the ga me \\ b u t r ar e ly breached t he Sy rian d ef enc e .\end{tabular} & 43 \\ \cdashline{2-3}
 & \begin{tabular}[c]{@{}l@{}}Ja pan t h e n la id siege t o the Syrian penal ty are a f or mos t of the gam e \\ but ra rely breached t he Sy r ian defence .\end{tabular} & 36 \\ \cdashline{2-3}
 & \begin{tabular}[c]{@{}l@{}}J ap an th en l aid s ie ge to th e Syrian pen a lt y ar ea for most o f t he game \\ b ut r are ly bre ache d th e Syrian de fen ce .\end{tabular} & 42 \\ \hline
K-means & \begin{tabular}[c]{@{}l@{}}Ja pan then laid siege to t he Sy r ian penalty are a for most of t he gam e \\ bu t r are ly breached the Syri an def enc e .\end{tabular} & 31 \\ \cdashline{2-3}
 & \begin{tabular}[c]{@{}l@{}}J ap an the n laid siege t o the Syrian pen al ty area for most of the game \\ but r ar e ly bre ac hed the Syrian defence .\end{tabular} & 36 \\ \cdashline{2-3}
 & \begin{tabular}[c]{@{}l@{}}Japan then laid si e ge to th e Syri an p en al t y area for most o f t h e game \\ but rarely breached the Sy ri an defence .\end{tabular} & 34 \\ \hline \hline
\end{tabular}
\caption{Examples of subword sequences by each selection method. Subword breaks are shown in the blanks.}
\label{tab:tokenization_examples}
\end{table*}

In \S \ref{sec:inference}, we use 3 subword sequence selection methods: Random, Cos-Sim, and K-means. We show specific examples of the subword sequences obtained by each selection method in Table~\ref{tab:tokenization_examples}.

\section{Detail About Dataset}
\label{sec:detail_dataset}
\subsection{NER}

\begin{figure*}[t]
  \includegraphics[width=\linewidth]{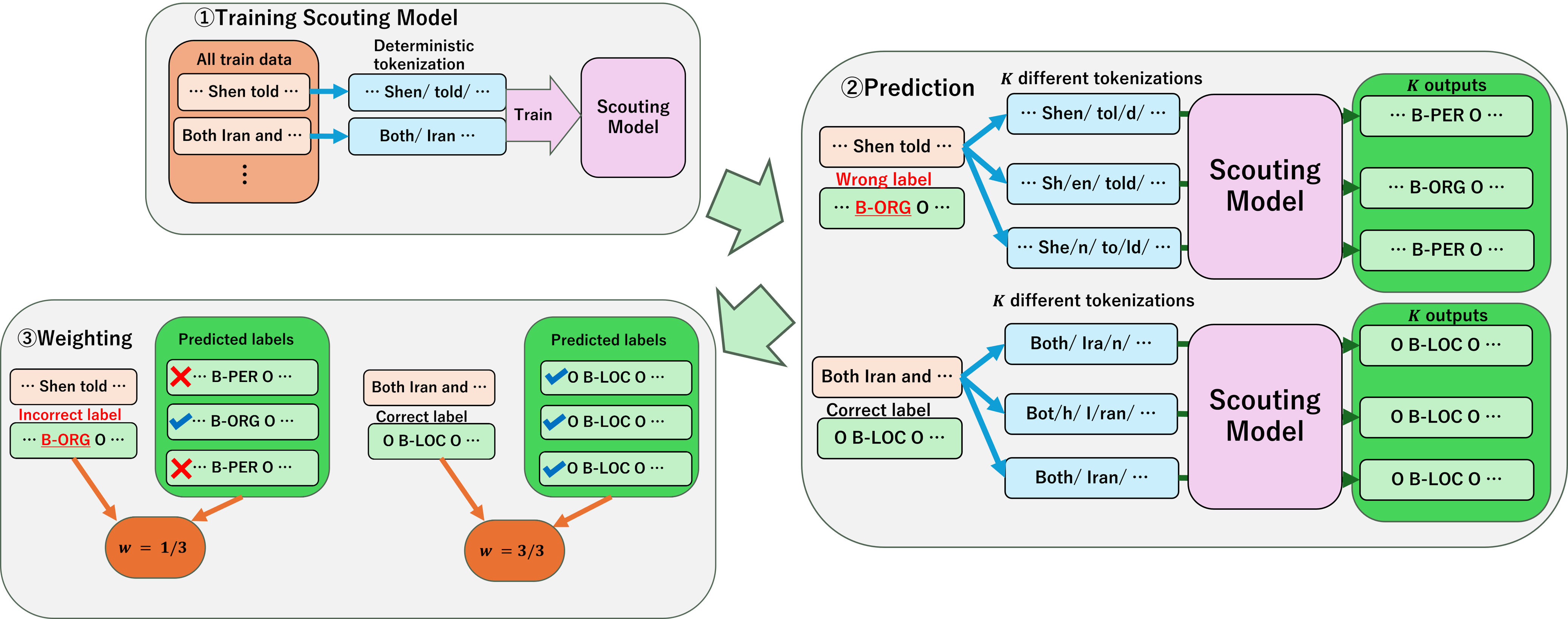}
  \caption {The overview of SubRegWeigh in the case of NER.}
  \label{fig:overview_ner}
\end{figure*}

The overview in the case of NER is shown in Figure~\ref{fig:overview_ner}.

We use the training split of \textbf{CoNLL-2003}~\cite{conll2003} as the target of the annotation weighing.
This split is also used to train the final NER models evaluated on the following test datasets.
The training split comprises 946 articles (14,041 sentences and 23,499 named entity labels).

We used the validation and test split of \textbf{CoNLL-2003} to evaluate the finally trained NER models.
The validation split contains 216 articles (3,250 sentences and 5,942 named entity labels) and the test split contains 232 articles (3,453 sentences and 5,648 labels).

In addition to the original CoNLL dataset, we employ the modified version and the recently published version of this dataset for further evaluation.
\textbf{CoNLL-CW}~\cite{CrossWeigh} is constructed by manually correcting annotation errors in the CoNLL-2003 test split, which includes 231 articles (3,453 sentences and 5648 named entity labels).
\textbf{CoNLL-2020}~\cite{conll2023} is constructed with articles from the year 2020 using the same definitions of NER labels as CoNLL-2003, which consists of 131 articles (1,840 sentences and 4,007 named entity labels).

\subsection{Text Classification}
We used the training and validation split of \textbf{SST-2}~\cite{sst2}.
The training split comprises 67,349 sentences. 55.8 \% of this split are negative labels, and the rest are positive labels. 
The validation split comprises 872 sentences. 50.9 \% of this split are negative labels, and the rest are positive labels. 
Test split is not used because the labels have not been published. 

\section{Detail About Model Setting}
\label{sec:detail_model_settings}

\begin{table}[t]
    \centering
    \small
    \begin{tabular}{lrrr}
    \hline \hline
    Model & \multicolumn{2}{c}{$\mathrm{RoBERTa}$} & \multicolumn{1}{c}{$\mathrm{LUKE}$} \\
    Task & \multicolumn{1}{c}{NER} & \multicolumn{1}{c}{\begin{tabular}[c]{@{}c@{}}Text \\ Classification\end{tabular}} & \multicolumn{1}{c}{NER} \\ \hline
    Epoch & 5(20)* & 5(20)* & 5 \\
    Learning rate & 1e-5 & 5e-5 & 1e-5 \\
    Batch size & 32 & 32 & 16 \\
    Weight decay & 0.01 & 0.01 & 0.01 \\
    Params & \begin{tabular}[c]{@{}r@{}}354M\\ (LARGE)\end{tabular} & \begin{tabular}[c]{@{}r@{}}354M\\ (LARGE)\end{tabular} & \begin{tabular}[c]{@{}r@{}}560M\\ (LARGE)\end{tabular} \\
    GPU & V100*1 & A100*1 & A100*1 \\  \hline \hline
    \end{tabular}
    \caption{Model hyperparameters. *In $\mathrm{RoBERTa}$, the scouting model was trained in 5 epochs and the final model was trained in 20 epochs. Other hyper-parameters are the same in the scouting and final model.}
    \label{tab:hyper_param}
\end{table}

We employed $\mathrm{RoBERTa_{LARGE}}$~\cite{roberta} and $\mathrm{LUKE_{LARGE}}$~\cite{luke} for the scouting model $M$ and the final model $M'$. 
The scouting and the final model were trained independently. 
Each model was trained using the hyperparameters in Table~\ref{tab:hyper_param}. 

We used the same architectures for the scouting and final models.
In other words, the scouting model with $\mathrm{RoBERTa_{LARGE}}$ is used to weigh the annotation errors that are used for the training of the final model with $\mathrm{RoBERTa_{LARGE}}$.
When the final model is based on $\mathrm{LUKE_{LARGE}}$, we used the scouting model with $\mathrm{LUKE_{LARGE}}$.

In NER, for $\mathrm{RoBERTa_{LARGE}}$, inference was performed with a token-level classification model where the first subword of each word was classified based on BIO tags, and the outputs of the second and subsequent subwords were masked and ignored. For $\mathrm{LUKE_{LARGE}}$, inference was performed using the span classification method employed in LUKE~\cite{luke}, which classifies whether a span from one word to another is a named entity.

\section{Pseudo-Incorrect Labeling Method}
\label{sec:pseudo_incorrect_label_creation}
In the experiments in \S \ref{sec:pseudo_incorrect_label_test}, Some of the original labels were replaced with pseudo-incorrect labels. This section describes the replacement method. We replaced the labels so that the replacement would not conflict with the IOB2 format and the number of entities would not increase too much from the original data. Specifically, in the case of replacing $a \%$ of all labels, we used the following procedure.
\begin{enumerate}
    \item Select one label at random.
    \item Select one label at random from B-xxx or O that is different from the selected label and replace that label with the selected label.
    \item Apply the following operations, depending on the labels to change and to be changed.
    \begin{enumerate}
        \item If changing from O to B-xxx and the label behind the selected label is B-xxx, change it to I-xxx.
        \item If changing from B-xxx or I-xxx to O and the label behind the selected label is I-xxx, change it to B-xxx.
        \item If changing from B-xxx or I-xxx to B-yyy, change the subsequent I-xxx to I-yyy. 
    \end{enumerate}
    \item Repeat until the number of changed labels reaches $a \%$ of all labels.  
\end{enumerate}

\section{Effect of Subword Regularization}
\label{sec:differenct_p}
\begin{table}[t]
\small
\centering
\begin{tabular}{rrrrrr}
\hline\hline
\multicolumn{1}{c}{\(p\)}  & \multicolumn{1}{c}{\(N\)} & \multicolumn{1}{c}{Method} & Time & \multicolumn{1}{c}{\begin{tabular}[c]{@{}c@{}}CoNLL\\ test\end{tabular}} & \multicolumn{1}{c}{\begin{tabular}[c]{@{}c@{}}CoNLL\\ CW\end{tabular}}\\ \hline
0.2 & 10 & Random & 3:26 & 93.68 & 95.28 \\
0.2 & 100 & Cos-Sim& 3:45 & 93.42 & 95.15 \\
0.2 & 100 & $K$-means& 3:58 & 93.52 & 95.20\\ \hdashline
0.1 & 10 & Random & 3:26 & 93.51 & 95.24\\
0.1 & 500 & Cos-Sim& 4:51 & 93.44 & 95.17\\
0.1 & 500 & $K$-means& 5:21 & 93.81 & 95.45\\ \hline \hline
\end{tabular}
\caption{Difference in the speed and performance against the hyperparameter of subword regularization $p$.}
\label{tab:p_and_n}
\end{table}
We select a few tokenization candidates from a large number of candidates to use a wide range of different candidates from the original tokenization for the annotation weighing. 
Instead of using a large number of candidates, we can also obtain various candidates by changing the BPE-Dropout hyperparameter $p$. Therefore, we investigated the difference in \(F_1\) score of the final model against $p$ in the NER dataset.
In this examination, we used \(p=0.2\) as the hyperparameter of BPE-Dropout, which tends to generate more different tokenization candidates than $p=0.1$. the number of tokenization candidates was limited to \(N=100\) to explore the possibility of more efficient annotation weighing by increasing the p of BPE-Dropout to generate diverse subword sequences.
For Random, the number of generated subword sequences \(N\) always equals the number of subword sequences used for inference \(K\). 
Since the experiment was conducted with \(K=10\), we set to \(N=10\) in the experiment for Random.

The results are shown in Table~\ref{tab:p_and_n}. 
One can see that the larger $p$ improves the performance with Random, while the weighing time did not change.
This suggests that Random with \(p=0.1\) did not obtain a sufficient variety of tokenization candidates. 
For non-Random selection methods, the time for weighing annotation errors was reduced to almost equivalent to the ones by Random.
However, \(F_1\) score decreased. The non-random selection methods use TF-IDF when comparing subword sequences. Therefore, when selecting from high $p$ and small $N$, the number of subwords that appear only in a single sequence will increase, and the selection will be attracted by such subwords. In addition, subwords that appear only in a single sequence are more finely segmented subwords, i.e., almost character-level subwords, so that a model trained with deterministic subwords cannot make correct inferences. This is likely the reason for the low $f_1$ scores of the non-Random selection methods with $p=0.2$.
From these results, it is evident that using a large \(N\) with a small \(p\) is appropriate for balancing both speed and performance.

\section{Effect of Minimum Weight}

\begin{table}[t]
\centering
\begin{tabular}{rrr}
\hline \hline
\multicolumn{1}{c}{$w_{min}$} &
  \multicolumn{1}{c}{\begin{tabular}[c]{@{}c@{}}CoNLL\\ test\end{tabular}} &
  \multicolumn{1}{c}{\begin{tabular}[c]{@{}c@{}}CoNLL\\ CW\end{tabular}} \\\hline
1 (Vanilla) & 93.54 & 95.27 \\
0.7           & 93.58 & 95.23 \\
0.3           & 93.81 & 95.45 \\
0             & 93.25 & 95.08 \\ \hline \hline
\end{tabular}
\caption{Difference in the performance against $w_{min}$}
\label{tab:w_min}
\end{table}

In the experiment shown in \S \ref{sec:experiment}, the minimum weight $w_{min}$ for SubRegWeigh was set to 1/3. Herein, we investigate the effect on the $F_1$ scores of the CoNLL-2003 test data and CoNLL-CW when $w_{min}$ is changed, using RoBERTa LARGE. We use $w_{min} = 1/3, 2/3$, as well as $w_{min}=0$, where no minimum weight is set. For comparison, Vanilla is recorded with $w_{min}=1$, which is equivalent to not performing any weight correction, as all data are given 1 as the weight.

The results are shown in Table~\ref{tab:w_min}. 
When $w_{min}=0$, the $F_1$ scores decrease compared to other settings. This is likely because the weight of some data becomes 0, reducing the amount of data used for training, and therefore, the model is not sufficiently trained. This is similar to the decrease in CrossWeigh~\cite{CrossWeigh} when data with low weights were not used instead of being weighted. When $w_{min}=2/3$, the $F_1$ scores also decrease compared to $w_{min}=1/3$. This is likely because the dataset is weighed between 2/3 and 1, making the dataset almost identical to the vanilla baseline. This result indicates that $w_{min}$ should be set to a low but non-zero value.

\section{Additional Datasets}

\begin{table}
\begin{tabular}{lrr}
\hline \hline
 & \multicolumn{1}{c}{WNUT2017} & \multicolumn{1}{c}{MRPC} \\
 & \multicolumn{1}{c}{\(F_1\)} & \multicolumn{1}{c}{ACC} \\ \hline
SoTA & $\mathbf{60.45}^{\dagger\dagger\dagger}$ & \multicolumn{1}{c}{-} \\ \hline
Vanilla & $60.04_{\pm0.31}$ & $\mathbf{90.43_{\pm0.15}}$ \\
CrossWeigh & $60.19_{\pm0.43}$ & $90.35_{\pm0.23}$ \\
SubRegWeigh & \multicolumn{1}{l}{} &  \\
\multicolumn{1}{r}{Random} & $60.15_{\pm0.30}$ & $90.16_{\pm0.23}$ \\
\multicolumn{1}{r}{Cos-Sim} & $60.05_{\pm0.46}$ & $89.69_{\pm0.36}$ \\
\multicolumn{1}{r}{$K$-means} & $60.29_{\pm0.41}$ & $86.82_{\pm0.45}$ \\ \hline\hline
\end{tabular}
\caption{The results of the additional datasets. ${}^{\dagger\dagger\dagger}$ is SoTA score from \newcite{clkl}.
}
\label{tab:additonal_datasets}
\end{table}

We experimented with additional datasets, WNUT2017~\cite{wnut17} and MRPC~\cite{mrpc}. 
We use the train and test split of WNUT2017 and the train and develop split of MRPC. 
The basic experiment setup is the same as for \S \ref{sec:model_settings} and Appendix~\ref{sec:detail_model_settings}, but in WNUT2017, we use only RoBERTa and the URLs in the text are replaced with <URL> tags.

The results are shown in  Table~\ref{tab:additonal_datasets}.
In WNUT2017, the proposed method improves for baselines as in the CoNLL2003 experiment. However, in MRPC, the proposed method has worse accuracy than baseline and CrossWeigh. MRPC is a task to classify whether two sentences have the same meaning. the same words are often used in each of these two sentences. If subword regularization splits these words into different subwords, the scouting model cannot perform inference successfully. This is the reason for the accuracy deterioration in MRPC.

\end{document}